\documentclass[10pt,twocolumn,letterpaper]{article}

\usepackage{cvpr}
\usepackage{times}
\usepackage{epsfig}
\usepackage{graphicx}
\usepackage{amsmath}
\usepackage{amssymb}

\usepackage[pagebackref=true,breaklinks=true,letterpaper=true,colorlinks,bookmarks=false]{hyperref}

\cvprfinalcopy 

\def\cvprPaperID{1746} 

\ifcvprfinal\fi
\begin{document}

\title{IF-TTN: Information Fused Temporal Transformation Network \\
	for Video Action Recognition}

\author{Ke Yang \quad Peng Qiao \quad Dongsheng Li \quad Yong Dou\\
National University of Defense Technology, China\\
}

\maketitle

\begin{abstract}
Effective spatiotemporal feature representation is crucial to the video-based 
action recognition task. Focusing on discriminate spatiotemporal feature 
learning, we propose Information Fused Temporal Transformation Network 
(IF-TTN) for 
action recognition on top of popular Temporal Segment Network (TSN) framework. 
In the network, Information Fusion Module (IFM) is designed to fuse the 
appearance and motion features at multiple ConvNet levels for each video 
snippet, forming a short-term video descriptor. With fused features as 
inputs, Temporal Transformation Networks (TTN) are employed to model
middle-term temporal transformation between the neighboring snippets following 
a sequential 
order. As TSN itself depicts long-term temporal 
structure by segmental consensus, the proposed network comprehensively 
considers 
multiple granularity temporal features.
Our IF-TTN achieves the state-of-the-art results on two most popular action 
recognition datasets: UCF101 and HMDB51. Empirical investigation reveals 
that our architecture is robust to the input motion map quality. Replacing 
optical flow with the motion vectors from compressed video stream, the 
performance is still comparable to the flow-based methods while the testing 
speed is 10x faster.
\end{abstract}

\section{Introduction}

Video action recognition has been widely studied by the 
computer 
vision community \cite{simonyan2014two,tran2015learning} as it can be 
applied in many areas like intelligent video surveillance and human behavior 
analysis. Since CNNs have achieved great successes in image classification task 
\cite{he2016deep,simonyan2015very,krizhevsky2012imagenet} and video action 
recognition can be considered as a classification task, a lot of CNN-based 
action recognition methods have been proposed 
\cite{tran2015learning,wang2016temporal,simonyan2014two,carreira2017quo}.
Compared to the image classification methods, temporal information is also 
critical for video action recognition. Appearances and dynamics are crucial and 
complementary aspects. The performance of video action recognition highly 
depends on how the algorithms utilize the relevant temporal information in 
cooperation with spatial features. Many CNN-based action recognition methods 
are proposed to classify videos in terms of their spatiotemporal features 
\cite{simonyan2014two,tran2015learning,wang2011action,wang2016temporal,carreira2017quo}.
Among them, Two-Stream CNN \cite{simonyan2014two} and C3D 
\cite{tran2015learning} are the most representative methods.
\begin{figure}[t]
	\centering
	\includegraphics[width=0.47\textwidth]{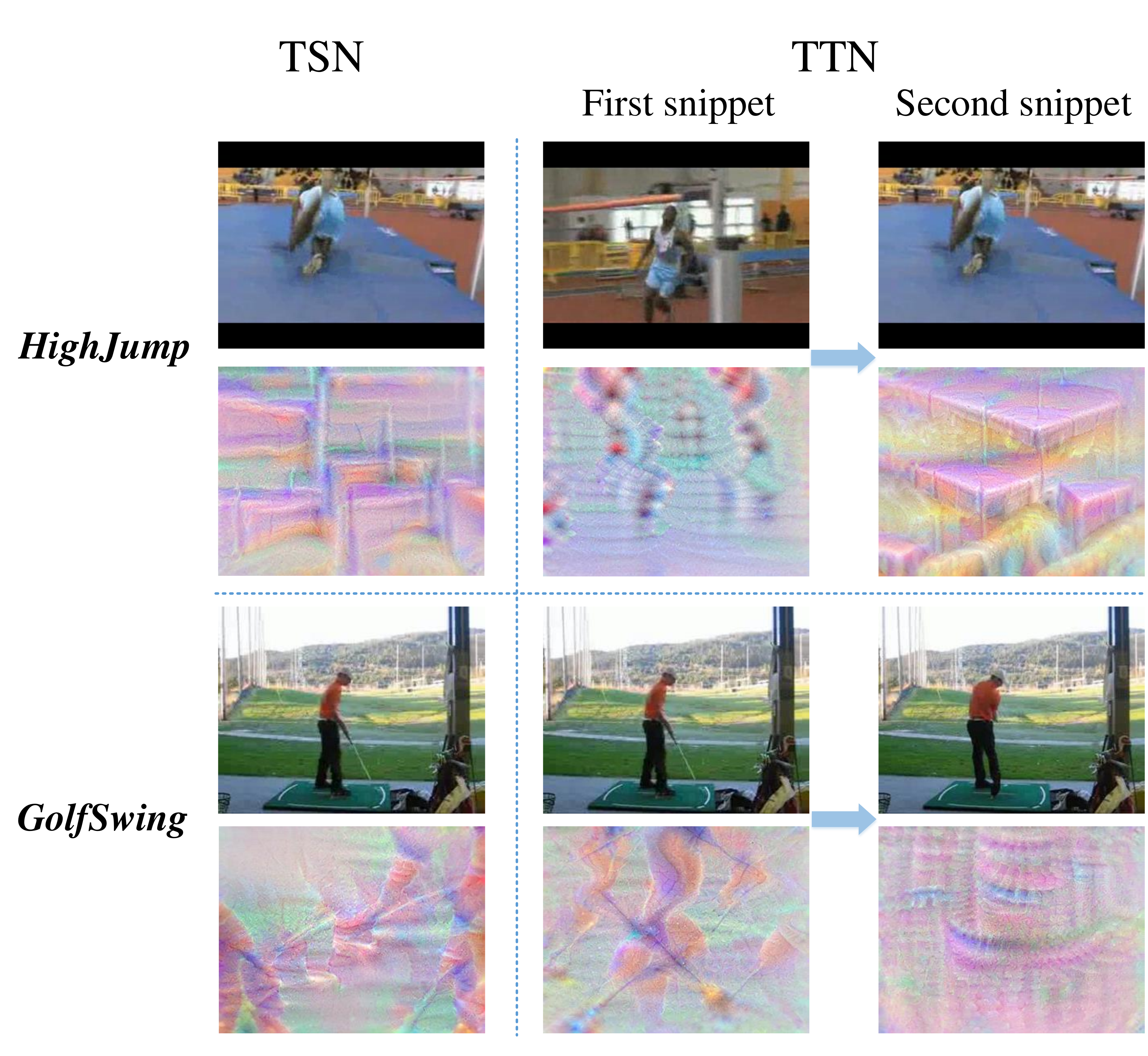}
	\caption{Class visualization of TSN and TTN models using DeepDraw on two 
		action categories:  ``\textit{HighJump}'' and ``\textit{GolfSwing}''.  
		For each category, visualized images are shown on bottom row, and the 
		RGB images similar to the visualized images are placed on top row. The 
		results of TSN  are shown in left column, and the results of TTN  
		are shown in middle and right columns since TTN takes snippets from two 
		adjacent segments as input. We can observe that TTN does capture the 
		temporal order between snippets  while TSN mainly replies on the object 
		and scene. 	
	}
	\label{fig_first}
\end{figure}

In a common Two-Stream CNN framework, appearances and dynamics are often 
decoupled and lost valuable connection during learning feature. 
Intuitively, human beings identify a specific action from video mainly by 
recognizing dynamics over appearances, namely, motions of objects rather than 
recognizing dynamics and appearances separately. C3D is proposed to encode 
appearance and motion information simultaneously by 3D convolution upon 
multiple consecutive video frames, but its performance is 
limited compared with Two-Stream based methods, which means that the effective 
fusion of spatial and temporal features is still under exploration.

Besides spatial and temporal feature fusion, temporal order modeling is also 
lack of studies. Most of the existing works rely on short-term temporal 
modeling due to small temporal receptive window size and consecutively 
sampling strategy. To model long-term temporal structure, Temporal Segment 
Network \cite{wang2016temporal} sparsely sampled frames and aggregated snippet 
features over a whole video. However, it simply treats a video as a bag of 
snippets and does not capture the temporal order that reflects transformations 
between video snippets.

In this paper, we propose Information Fused Temporal Transformation Network 
(IF-TTN) for video action recognition based on the Temporal Segment Network 
(TSN) framework. In order to extract more effective spatiotemporal features, we 
proposed an Information Fusion Module (IFM) to fuse the appearance and motion 
features at multiple ConvNet levels for each video snippet. The fused features 
depict what (captured by the spatial stream) moves in which way (captured by 
the temporal stream). With the fused features, we designed a Temporal 
Transformation Network (TTN) to model the temporal order between the 
neighboring snippets. 

The visualization results of TTN are shown in Figure \ref{fig_first}. It can be 
observed that TTN does learn the transformation between neighboring snippets. 
Taking the ``\textit{HighJump}'' in Figure \ref{fig_first} as an example, TTN 
models the transformation between human running in front of a high jump 
crossbar and human falling on the mat after skipping the crossbar. The 
generated image of the first snippet depicts the running human. The generated 
image of the second snippet depicts the mat that the jumper falling on. 
Convolution is invariant to translation and scale, thus objects in 
generated images, such as people and mat, are with different scales and spatial 
locations, which makes the generated images look cluttered. 

In addition, 
this kind of temporal transformations between snippets actually describe the 
mid-term temporal structure of a video, which is complementary to the 
short-term temporal descriptor and long-term temporal structure. Therefore, our 
network comprehensively considers multiple granularity temporal features.
In our work, the reasonably structured modeling of the temporal features and 
the complementary fusion from multi-level spatial features reduce the 
dependency on motion input quality. Replacing optical flow with the motion 
vectors from compressed video stream, the performance is still competitive with 
the optical flow-based methods while the testing speed is 10x faster. Our 
contributions can be summarized as follows:

\begin{itemize}
	
	\item We design IFM to fuse two stream features. This design 
	involves appearance and motion information simultaneously and further 
	benefits the 
	temporal modeling.
	\item We design TTN to model the temporal order of video 
	snippets at multiple feature levels. 
	
	\item We combine IFM and TTN to form IF-TTN, which can be trained in an 
	end-to-end manner. IF-TTN is robust to motion input quality 
	owing to effective feature 
	learning, which makes it practical in the real-time scenarios. 
	IF-TTN achieves	state-of-the-art results on both non-real-time and 
	real-time action recognition tasks.
\end{itemize}

\begin{figure*}[t]
	\centering
	\includegraphics[width=1\textwidth]{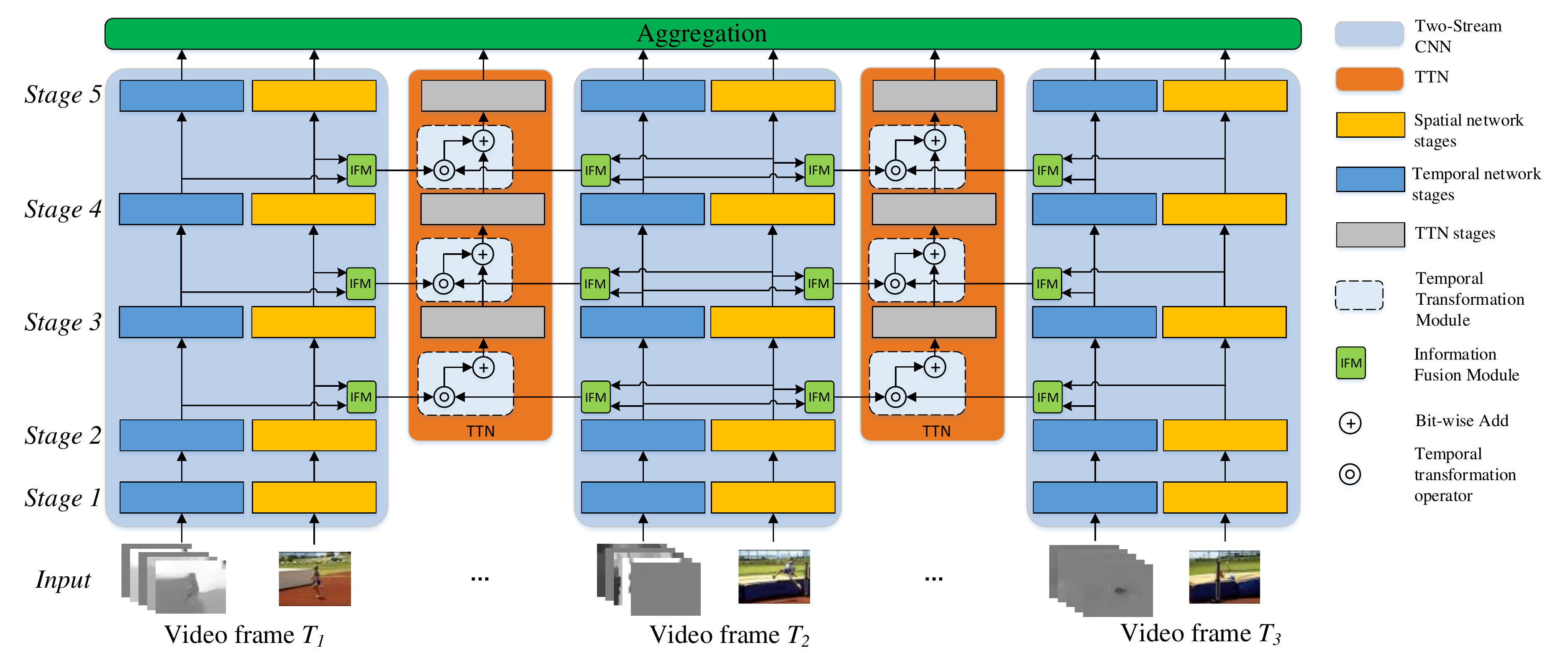}
	\caption{Overall architecture  of IF-TTN.  A video is divided into $K$ 
		segments ($K$ is set to 3 in this illustration). From each segment a 
		video frame $T_k$ is randomly sampled to represent corresponding 
		segment. 
		These 
		frames are  arranged in a strict temporal order
		and passed through Two-Stream CNN to extract appearance 
		and temporal features at multiple network stages.
		Two stream features are then fused	by IFM and fed to TTN. TTN takes 
		features from previous and next sampled video frames as input and  
		models the 
		temporal order between them.}
	\label{fig_overall}
\end{figure*}
\section{Related Work}
\label{re_work}
\textbf{Action recognition:} Improved Dense Trajectory Feature (iDTF) 
\cite{wang2011action,wang2013action} has been in a dominant position in the 
field of action recognition. Recently, 2D Convolutional Neural Networks trained 
on ImageNet \cite{russakovsky2015} were employed to perform RGB image 
classification. But their performance on video classification was limited as 
they can only capture appearance information. In order to model motion 
information, Two-Stream CNN was proposed and got a significantly boost in 
performance by taking both RGB images and optical flow as inputs. To model 
spatiotemporal feature better, Tran \textit{et al.} proposed 3D CNN 
architecture called C3D in an attempt to directly extract high-level semantics 
spatiotemporal abstraction from raw videos \cite{tran2015learning} and then 
proposed Res3D to further improve recognition performance 
\cite{tran2017convnet}. To take advantage of both Two-Stream CNN and 3D CNN, a 
Two-Stream Inflated 3D CNN (I3D) was proposed and allowed for initialization 
with ImageNet pre-trained weights \cite{carreira2017quo}. 


\textbf{Temporal Structure Modeling:}
Plenty of works have been dedicated to model the temporal structure for action 
recognition 
\cite{niebles2010modeling,gaidon2013temporal,wang2014video,wang2016temporal}. 
With the development of Deep Learning, many recent works modeled the temporal 
structure via network design.
Temporal Segment Network (TSN) \cite{wang2016temporal} was proposed to model 
temporal structure on the entire videos in an end-to-end manner. However, TSN 
failed to capture the temporal order of video frames. Zhou \textit{et al.} 
proposed a Temporal Relation Network (TRN) \cite{zhou2017temporal}  to learn 
and reason about temporal dependencies between video frames at multiple time 
scales. In \cite{yue2015beyond} and \cite{donahue2015long}, Long Short-Term 
Memory (LSTM) networks were used to capture the long-range dynamics for action 
recognition.

%
%

\textbf{Real-time action recognition:}
State-of-the-art video understanding methods relied heavily on optical flow. 
The heavy computation cost of optical flow prevented these methods from 
real-time implementation. There were a few works dealt with real-time video 
understanding by replacing the costly optical flow with low-cost motion 
representations. Bilen \textit{et al.} proposed dynamic image (DI) 
\cite{bilen2016dynamic} to simulate the motion information, and Sun \textit{et 
	al.} proposed Optical Flow guided Feature (OFF) \cite{sun2017optical} to 
	model 
short-term temporal variation (e.g. at a temporal length of about 7 
frames\footnote{7 frames are calculated from the training strategy of 
	OFF. TSN Two-Stream 
	CNN used 5 stacked optical flow frames to model short-term motions. Thus 7 
	frames belong to short-term motions.}). 
Motion Vector (MV) was a coarse representation of motion, but it can be 
obtained 
directly from compressed video 
streams without extra calculation. Therefore, Enhanced Motion Vectors CNN 
(EMV-CNN) \cite{zhang2016real} used motion vector as the input of temporal CNN 
to 
improve inference speed and CoViAR \cite{wu2017compressed} adopted an 
accumulated 
motion vector for real-time action recognition. Suffered from the lack of fine 
detailed motion information in MV, recognition performance was degraded 
dramatically. The performance of both EMV-CNN and CoViAR was far behind 
Two-Stream CNN with optical flow.

\textbf{The works most similar to our work} are 
\cite{feichtenhofer2016spatiotemporal,feichtenhofer2017spatiotemporal} and 
\cite{zhang2016real}. The work in 
\cite{feichtenhofer2016spatiotemporal} studied the additive 
fusion 
of spatial and temporal features of Two-Stream CNN. Their follow-up work 
\cite{feichtenhofer2017spatiotemporal} studied the multiplicative 
fusion. Compared with that, our 
contribution is to design a more general and effective fusion module, which 
jointly operates additive and multiplicative interactions. Moreover, we use 
adaptively weighted interaction items during fusion and balance their impacts 
through learning on weight parameters. Experiment results also show that our 
fusion module performs better than the fusion type used in 
\cite{feichtenhofer2016spatiotemporal,feichtenhofer2017spatiotemporal} as 
reported in the Section \ref{Eval_exp}.
 
EMV-CNN \cite{zhang2016real} first used motion vectors as motion 
representation for real-time action recognition. They replaced the optical flow 
with motion vectors in 
Two-Stream CNN and developed transferring techniques to enhance the MV-CNN, but 
the performance was much lower than the state-of-the-art 
optical-flow-based methods. In this paper, we experimentally prove that 
motion-vector-based Two-Stream CNN 
can achieve comparable performance to optical-flow-based methods if we adopt 
effective feature learning rather than the simple usage of motion vectors. We 
build a model with more reasonably structured modeling and the 
complementary feature fusion to make the network tolerant to the low quality of 
motion input. Experiment results prove that our network is highly tolerant to 
the quality of motion input thanks to the combination of short-term 
spatiotemporal feature fusion, sequentially middle-term temporal modeling and 
long-term temporal consensus.

\section{Information Fusion Temporal Transformation Network}
\label{IFTRN}

In this section, we describe the Information Fusion module (IFM) that 
fuses the features from Two-Stream CNN, the Temporal 
Transformation Network (TTN) and the real-time adaption 
of our network.


The overall network architecture of IFM-TTN is shown in Figure 
\ref{fig_overall}. Given a video $V$ containing $N$ frames, we first divide 
it into $K$ segments $\{S_1,S_2,...,S_K\}$. For $k$-$th$ segment $S_k$, we 
randomly 
sample a frame $T_k$ from it, called snippet. The assemble of sampled frames 
${\{T_{k}\}}_{k=1}^K$ are 
fed into deep feature extractor for feature extraction. 
Each snippet is processed by Two-Stream CNN to 
extract features as $A_{k} = \phi_{s}(T_k;W_s)$ and $B_{k} = 
\phi_{t}(T_k;W_t)$. 
$A_{k}$ and $B_{k}$ are features extracted from spatial and temporal stream 
networks, 
respectively.  $\phi_{s}(W_s)$ and $\phi_{t}(W_t)$ are the functions 
representing the spatial stream CNN and temporal stream CNN with parameters 
$W_s$ and $W_t$, respectively.

Instead of using the features only from final convolutional layer, we 
involve the features from multiple stages of CNN to encode snippets at 
multiple spatial scales. 
Assuming that the CNN has $L$ stages, $A_{k} =\{ {a_{k}^l}\}_{k=l_s}^L$ and 
$B_{k} =\{ 
{b_{k}^l}\}_{k=l_s}^L$, where $l_s$ represents the stage from which we start to 
extract features.
\subsection{Information fusion module}\label{IFM}


\begin{figure}[t]
	\centering
	\includegraphics[width=0.4\textwidth]{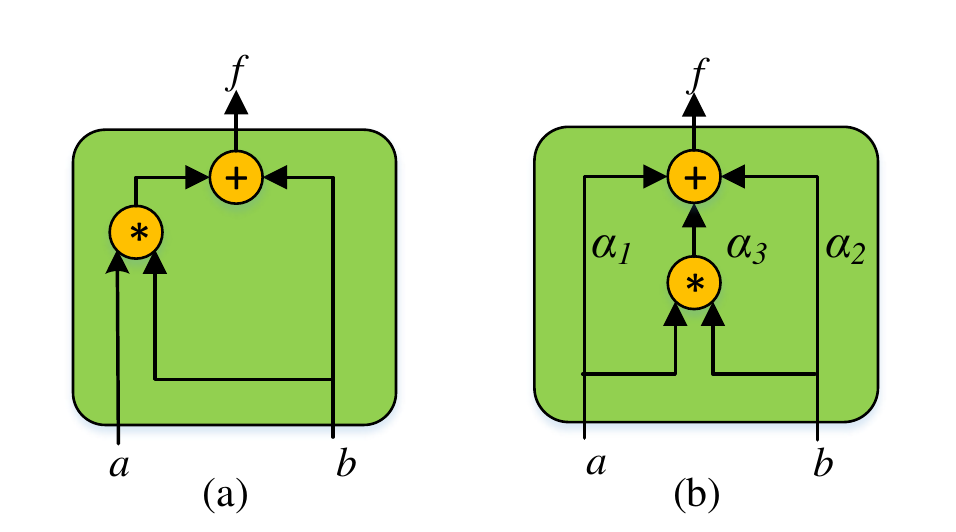}
	\caption{Illustration of information fusion module. (a) Attention based. 
		(b) Adaptive fusion.}
	\label{fig_ifm}
\end{figure}

\textbf{Feature fusion:}
It is desired to fuse the 
features of spatial and temporal networks to generate an efficient and compact 
representation for each snippet. Given a feature pair
$ a_{k}^l $ and $ b_{k}^l $, we 
can get 
the fused features with an Information Fusion Module (IFM):
\begin{equation}\label{eq:FM}
{ f_{k}^l  = {\mathcal{F}}(a_{k}^l, b_{k}^l) }
\end{equation}
\begin{figure*}[t]
	\centering
	\includegraphics[width=1.\textwidth]{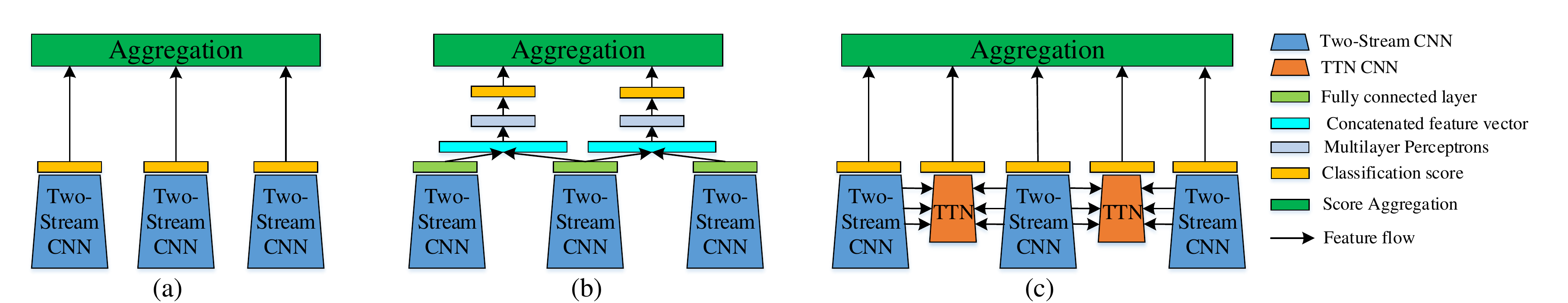}
	\caption{Illustration of temporal structure modeling. We only show 3 
		segments for 
		convenience. (a) Temporal Segment Network. (b) Temporal relation 
		Network (TRN) in 
		\cite{zhou2017temporal} . (c) Our Temporal Transformation Network 
		(TTN).}
	\label{fig_trn}
\end{figure*}

\noindent  where $\mathcal{F}$ represents the fusion function, $f_{k}^l$ 
denotes 
the fused 
features for $k$-$th$ segment at $l$-$th$ CNN stage . 

We investigate two implementations of fusion modules:

(1) \textbf{Attention based fusion module:} There exists a common insight that 
the temporal feature maps can act as the 
attention maps to the corresponding spatial feature maps, because optical flow 
can locate human foreground areas and is invariant to appearance. Besides the 
motion patterns, the scenes and objects in spatial stream are also important 
for classification, especially for the actions with subtle motions. For 
example, the recognition of musical instrument is important to the recognition 
of playing musical instrument. Taking account of these two factors, we 
formulate the function as:
\begin{equation}\label{eq:FM_Att}
{ f_{k}^l  =a_{k}^l + a_{k}^l \odot b_{k}^l }
\end{equation}

\noindent where $\odot$ corresponds to element-wise multiplication. 
$ a_{k}^l \odot b_{k}^l $ can also be viewed as a residual term. In other 
words, we enhance the features of interest rather than removing the features 
that are not attended.

(2) \textbf{Adaptive fusion module:} 
Recently, there are a series of works 
\cite{feichtenhofer2017spatiotemporal,feichtenhofer2016spatiotemporal} studying 
the fusion of spatial and temporal streams of Two-Stream CNN. In 
these works, additive \cite{feichtenhofer2016spatiotemporal} and multiplicative 
interactions \cite{feichtenhofer2017spatiotemporal} were considered separately. 
We propose adaptive fusion that covers interactions on both additive scale and 
multiplicative scale. We weight interaction items during fusion and balance 
their impacts through learning on weight parameters:
%
\begin{equation}\label{eq:FM_Adp}
{ f_{k}^l  = \alpha_1 \ast a_{k}^l + \alpha_2 \ast b_{k}^l + \alpha_3 \ast 
	a_{k}^l \odot 
	b_{k}^l }
\end{equation}

\noindent where $ \alpha_1 ,\alpha_2,\alpha_3$ are learnable weight parameters 
that 
update with the whole network. From Equation \ref{eq:FM_Att} and 
\ref{eq:FM_Adp}, it can be derived that attention based fusion module is a 
special case of adaptive fusion module when $\alpha_1 =1,\alpha_2 = 0,\alpha_3 
= 1$.

After fusion, the fused features $\{ {f_{k}}\}_{k=1}^K$ are ready to be fed to 
the TTN, where $f_{k}=\{f_k^l\}_{l=l_1}^L$ represent the fused features of 
$k$-$th$ snippet.

\subsection{Temporal transformation network}
\label{TTN}
Given a sequence of fused features $\{ {f_{k}}\}_{k=1}^K$, Temporal 
Transformation 
Network (TTN) is proposed to to model the 
pairwise temporal transformations as below:
\begin{equation}\label{eq:Relation}
{T(V)} = \sum_{i<j} \mathcal{R} (f_{i},f_{j} ; W_{ttn})
\end{equation}

\noindent where $T(V)$ denotes the TTN features over the whole video. 
$\mathcal{R}(W_{ttn})$ are the transformation function representing the TTN 
with network parameters $W_{ttn}$. TTN integrates the fused features of ordered 
snippet-pairs.

We construct TTN using a standard CNN architecture and take snippet features 
from multiple stages as inputs, as shown in Figure \ref{fig_overall}. In this 
way, low-level 
detailed features are kept while exploring the temporal relation. In order to 
simplify the network structure, we only keep the features pairs $f_i$ and  
$f_j$ from adjacent segments, that is, $j=i+1$.

\begin{figure}[t]
	\centering
	\includegraphics[width=0.45\textwidth]{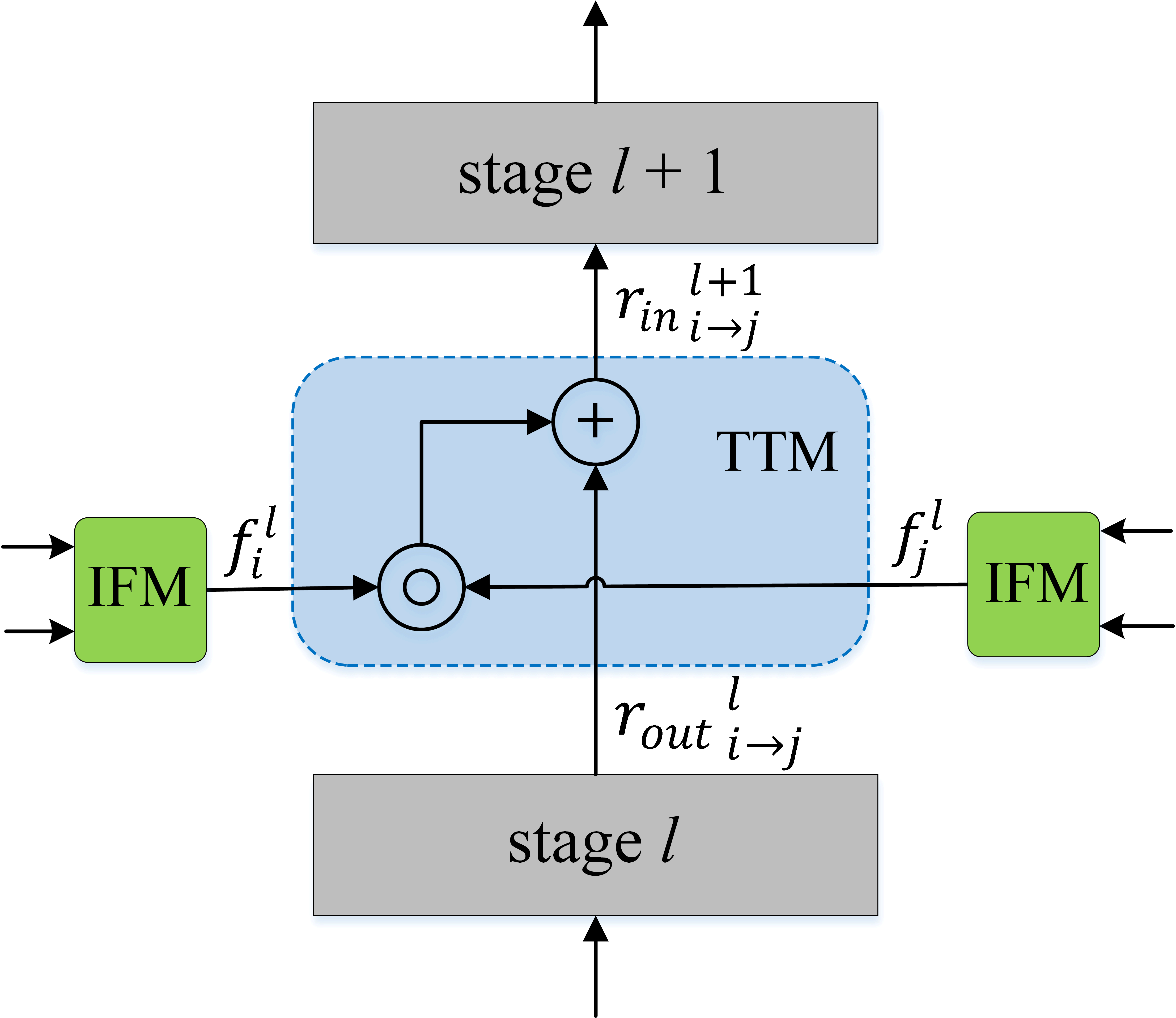}
	\caption{Illustration of TTM. This figure shows the 
		TTM between network stage $ l $ and stage $ l+1 $}
	\label{fig_merge}
\end{figure}

Between every two stages of the TTN, Temporal Transformation Modules (TTM) are 
designed to merge the features from adjacent segments. Figure \ref{fig_merge} 
shows the data flow of TTM, and the merging process can 
be formalized as follows:
\begin{equation}\label{eq:TRM}
{{r_{in\ }}_{i\rightarrow j}^{l+1}  =f_{i}^l \circledcirc f_{j}^l + 
	{r_{out\ }}_{i\rightarrow 
		j}^{l}}
\end{equation}

\noindent where $l$ represents stage index, $i, j$ represent segment index.  
$f_{i}^l, 
f_j^l$ are $l$-$th$ stage features from the $i, j$-$th$ segment, respectively. 
${r_{in\ }}_{i\rightarrow j}^{l+1}$ represents the input of $(l+1)$-$th$ stage 
in TTN, while ${r_{out\ }}_{i\rightarrow j}^{l}$ is the output features 
at $l$-$th$ stage.
$ \circledcirc $ denotes temporal transformation operator that modeling the 
temporal order of video segments. Image differences are commonly utilized to 
model the appearance change. Inspired by this, we use feature difference to 
reflect the ordered temporal transformation, that is, $ f_{i}^l \circledcirc 
f_{j}^l = f_{j}^l - f_{i}^l $. In this manner, the shape of feature map does 
not change, thus it allows us to use the pre-trained weights for TTN.

Similar to our TTN, Zhou \textit{et al.} \cite{zhou2017temporal} proposed 
a Temporal Relational Network (TRN) built on top of TSN to model 
the pairwise temporal relations between ordered frames. However, they only 
used the features extracted from the last fully connected layer  of CNN and 
deployed simple multilayer perceptrons (MLP) to model the relations. As the 
result, the spatial and low-level detailed features 
were completely lost before features are fed to the relation network.
One can easily find the differences among TSN, TRN, and TTN from 
Figure \ref{fig_trn}.
\subsection{Real-time adaption}\label{Real_time}

The networks based on Two-Stream CNN have achieved superior 
performance on recognition 
accuracy. However, the computation costs of optical flow make it impossible to 
apply these networks to the real scenarios. One popular solution is to use 
alternative motion representations as temporal stream CNN input, which could 
improve inference speed but might lead to degradation on recognition accuracy. 

Considering that motion vectors are inherently correlated with optical flow and 
can be extracted from compressed video stream directly with slight cost, it is 
desired to see whether the reasonable structured modeling and the complementary 
feature fusion  make IF-TTN tolerant to the low image quality of motion vectors.


We directly replace the input of temporal stream network with motion vectors. 
Before training motion-vector-based IF-TTN, we first train the 
optical-flow-based IF-TTN, and then initialize the motion-vector-based network 
with optical 
flow pre-trained weights. In our implementation, we do not use any image 
preprocessing techniques to improve the quality of motion vectors.

Since the extraction overhead of motion vectors is negligible, the video 
inference can be conducted at a very fast speed with a custom GPU. The adapted 
IF-TTN can be processed in real-time. 

Motion-vector-based CNN networks have been proposed in 
\cite{zhang2016real}, which simply replaced optical flow with motion vectors 
and transferred knowledge from optical-flow-based networks. 
Without in-depth 
exploration of the spatiotemporal structure, its performance was much lower 
than the state-of-the-art optical-flow-based methods. Our paper proves that 
motion-vector-based networks can achieve comparable performance to 
optical-flow-based networks with effective spatiotemporal feature modeling.
%
%
%


\subsection{Training and inference}
\label{TTN}

\textbf{Training:} Action recognition is a multi-class classification problem. 
We use the 
standard categorical cross-entropy loss to supervise the network optimization. 
In order to reduce the difficulty of 
training, we adopt a progressive multi-stage training strategy. First, we train 
a standard TSN \cite{wang2016temporal} with ResNet-50 backbone. Then, we freeze 
the TSN feature extractor, and train the TTN following a similar training 
strategy 
with TSN. Finally, we tune all the network jointly.

For the sake of better initialization for temporal network and TTN, 
following the 
good practice in \cite{wang2016temporal}, we first train the spatial network 
with ImageNet pre-trained weights. Then, we initialize the temporal network and 
TTN with pre-trained spatial network weights. This initialization method can 
speed up the training process and reduce the effect of over-fitting.

\textbf{Final predictions:} As there are
multiple classification scores produced by each segment, we first fuse 
the score of each stream network separately by averaging
the scores of all segments.
Then, we fuse the scores from Two-Stream CNN and TTN for final predictions.

\section{Experiments}\label{Eval}
In this section, we first introduce the evaluation datasets and the 
implementation details of our approach. Then we explore the contributions of 
each proposed module by the ablation experiments. Finally, we compare the 
performance of our method with the-state-of-the-art methods.

\subsection{Dataset}\label{Eval_dataset}

We evaluate our method on two popular video action recognition datasets: 
UCF-101 \cite{soomro2012ucf101} and HMDB-51 \cite{kuehne2011hmdb}. The UCF-101 
dataset contains 101 action classes and 13320
video clips, and the HMDB-51 dataset contains 6766 video clips from 51 action 
categories. Our experiments follow the official evaluation scheme which 
divides a dataset into three training and testing splits and report average 
accuracy over these three splits. For optical flow extraction, we use TVL1 
algorithm \cite{zach2007duality} implemented in OpenCV with CUDA. For motion 
vectors extraction, we use modified ffmpeg to extract motion vectors directly 
from 
compressed video stream without extra calculation.

\subsection{Implementation details} \label{Im_details}
We use ResNet-50 \cite{he2016deep} as our TSN backbone for both temporal and 
spatial streams. Our TTN is truncated from ResNet-50 and consists of three 
stages, namely stage 3, 4, 5 of ResNet-50. TTN does not involve stages lower 
than stage 3, as the fusion of the lower stages might suffer from noises and 
extreme large feature distances. Network truncation can greatly reduce 
computation cost and the consumption of GPU memory when training. For the 
segment 
number $K$, we set it to 7 to model the temporal structure.The average segment 
interval is around 1 second, which is closed to the length of an atomic action 
\cite{gu2017ava}. Therefore, the transformation between adjacent segments 
can be regarded as a mid-term motion. The Two-Stream CNN captures the temporal 
structure at a time length about 0.2 second which can be regarded as sub-atomic 
action or a short-term motion.

We use the mini-batch stochastic gradient descent (SGD) algorithm to optimize 
the network parameters. For spatial network, we initialize network weights with 
pre-trained models from ImageNet. Batch size is set to 64 and momentum set to 
0.9. Learning rate is initialized as 0.001 and decreases to its 0.1 every 30 
epochs. The maximum epoch is set as 80. After training spatial network, we 
initialize temporal network and TTN with pre-trained spatial network weights.  
For temporal network, we initialize the learning rate as 0.001, which reduces 
to its 0.1 every 100 epochs. The maximum epoch is set as 250. For TTN, we 
initialize the learning rate as 0.001, which reduces 
to its 0.1 every 80 epochs.The maximum epoch is set as 200. 

To alleviate over-fitting, we use strong data augmentation strategies and large 
drop ratios. For data augmentation techniques, we mainly follow
\cite{wang2016temporal} to do  location jittering, horizontal flipping, corner 
cropping, and scale jittering. The dropout ratio is set to 0.8 for spatial 
network and TTN while 0.7 for temporal network.

\begin{figure}[t]
	\centering
	\includegraphics[width=0.48\textwidth]{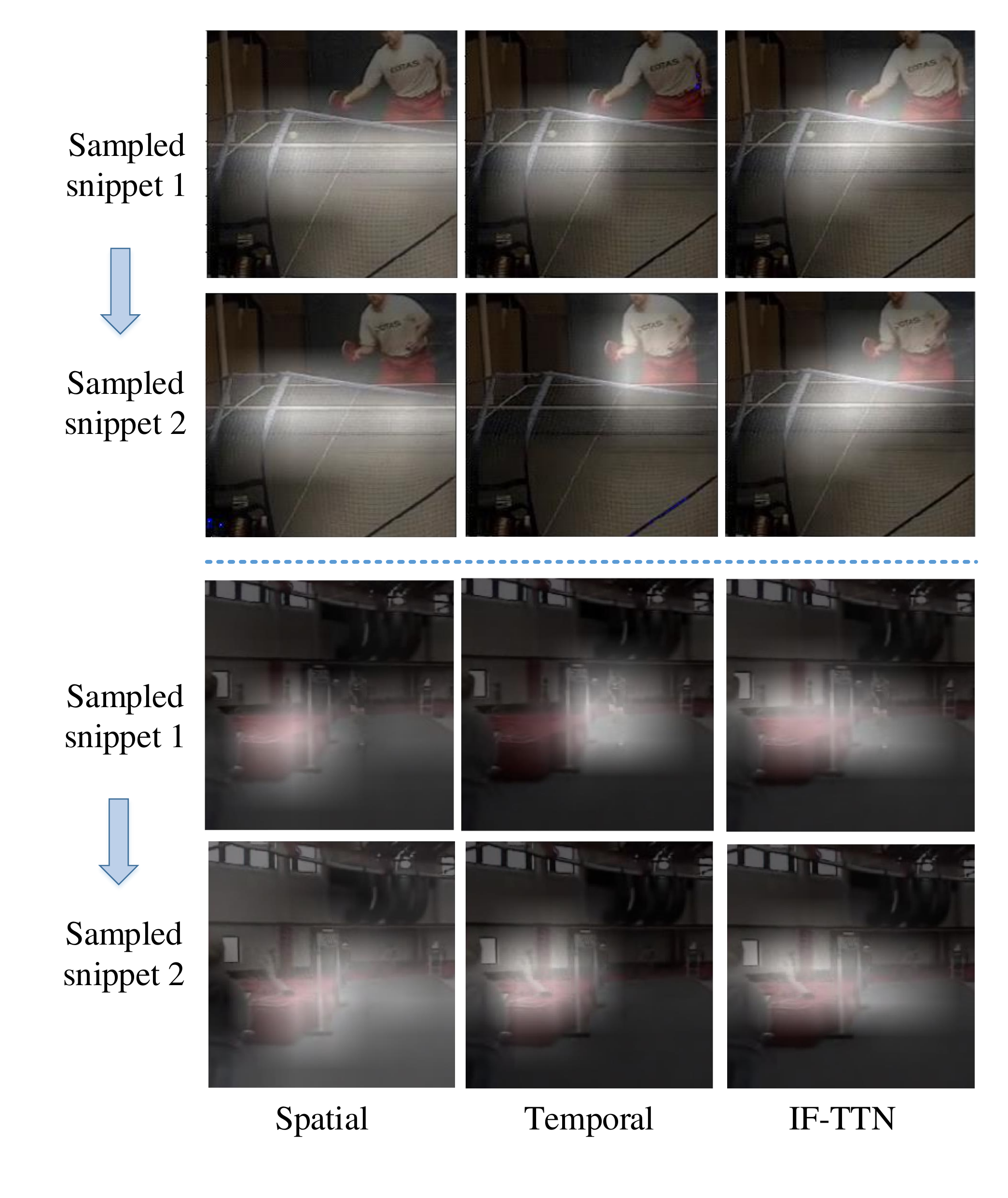}
	\caption{CAM visualization of spatial network, temporal network and IF-TTN.
		The first sampled snippet is shown in first row and the second in 
		second 
		row. Since IF-TTN takes two snippets as input, thus the CAMs of two 
		snippets are the same.}
	\label{fig_vis_ifm}
\end{figure}

\subsection{Exploration Study}\label{Eval_exp}

In this part, we study the contributions of each module of our approach.
All exploration studies are performed on UCF-101 dataset.

\textbf{Study on IFM:}
We propose IFM to fuse appearance and motion features for each video snippet at 
multiple ConvNet levels. To verify the effect of IFM, we conduct experiments 
under two settings: (1) The spatial and temporal streams are processed 
separately and TTNs are applied to two streams respectively; (2) IFM are used 
to 
fuse the features from Two-Stream CNN and TTN is applied to fused features. All 
other settings are set to the same. The experimental results are summarized in 
Table \ref{table_IFM}. From the results,  the attention based fusion and 
adaptive fusion both 
significantly improve performance. We attribute the improvements to the ability 
of IFM to model better spatiotemporal features of a short video snippet. Since 
two types of IFM achieve equal performance, we use attention based IFM in the 
following experiments for simplicity. It is worth noting that our IFM-TTN only 
has three CNNs while separate Two-Stream TTN has four CNNs, because both 
spatial and temporal networks have a TTN. Therefore, IF-TTN performs much 
better while has much less parameters.
\begin{table}[t]
	\caption{Comparison of experimental results whether using IFM modules or 
		not. Experiments are conducted on UCF101 split 1.}
	\label{table_IFM}
	\footnotesize
	\centering
	\begin{tabular}{c|ccccc|c|ccccc}
		\multicolumn{1}{c|}{Method} & \multicolumn{1}{c}{Accuracy (\%)}\\
		\hline
		\multicolumn{1}{c|}{Separate Two-Stream} & 
		\multicolumn{1}{c}{94.0}\\
		\hline
		\multicolumn{1}{c|}{Attention IFM Two-Stream} & 
		\multicolumn{1}{c}{95.0}\\
		\hline
		\multicolumn{1}{c|}{Adaptive IFM Two-Stream} & 
		\multicolumn{1}{c}{95.0}\\
		
	\end{tabular}
\end{table}

\begin{table}[t]
	\caption{Comparison of experimental results whether using different fusion 
		types. Experiments are conducted on UCF101 split 1.}
	\label{table_IFM_com}
	\footnotesize
	\centering
	\begin{tabular}{c|ccccc|c|ccccc}
		\multicolumn{1}{c|}{Method} & \multicolumn{1}{c}{Accuracy (\%)}\\
		\hline
		\multicolumn{1}{c|}{additive fusion 
			\cite{feichtenhofer2016spatiotemporal}} & 
		\multicolumn{1}{c}{93.8}\\
		\hline
		\multicolumn{1}{c|}{multiplicative fusion 
			\cite{feichtenhofer2017spatiotemporal}} & 
		\multicolumn{1}{c}{94.0}\\
		\hline
		\multicolumn{1}{c|}{Our IFM} & 
		\multicolumn{1}{c}{95.0}\\
		
	\end{tabular}
\end{table}

\begin{table}[t]
	\caption{Ablation study of IF-TTN. Experiments are conducted on UCF101 
		split 1.}
	\label{table_TTN}
	\footnotesize
	\centering
	\begin{center}
		\begin{tabular}{c|c}
			\multicolumn{1}{c|}{Method} & \multicolumn{1}{c}{acc.(\%)}\\
			\hline
			\multicolumn{1}{c|}{Spatial stream CNN} & 
			\multicolumn{1}{c}{84.9} \\		
			\multicolumn{1}{c|}{Temporal stream CNN} & 
			\multicolumn{1}{c}{86.9} \\
			\multicolumn{1}{c|}{Two-stream CNN} & 
			\multicolumn{1}{c}{93.1} \\
			\multicolumn{1}{c|}{TTN branch} & 
			\multicolumn{1}{c}{92.3} \\
			\hline
			\multicolumn{1}{c|}{complete IF-TTN} & 
			\multicolumn{1}{c}{95.0} \\
			
			\hline
		\end{tabular}
	\end{center}
	
\end{table}

\begin{table}[t]
	\caption{Experimental study of motion input study. Experiments are 
		conducted on UCF101 first split.}
	\label{table_MV}
	\footnotesize
	\centering
	\begin{tabular}{c|ccccc|c|ccccc}
		\multicolumn{1}{c|}{Method} & \multicolumn{1}{c}{optical flow} & 
		\multicolumn{1}{c}{motion vectors} &  \\
		\hline
		\multicolumn{1}{c|}{Spatial stream CNN} & 
		\multicolumn{1}{c}{84.9}  &
		\multicolumn{1}{c}{84.9} \\
		\multicolumn{1}{c|}{Temporal stream CNN} & 
		\multicolumn{1}{c}{86.8} & 
		\multicolumn{1}{c}{82.5} \\
		\hline
		\multicolumn{1}{c|}{IF-TTN} & 
		\multicolumn{1}{c}{95.0}& 
		\multicolumn{1}{c}{94.4}  \\
		
	\end{tabular}
\end{table}

We also perform comparative experiments to prove whether our IFM performs 
better than the fusion modules in 
\cite{feichtenhofer2016spatiotemporal,feichtenhofer2017spatiotemporal}. The 
work in \cite{feichtenhofer2016spatiotemporal} studied the additive 
fusion of spatial and temporal features of Two-Stream CNN. Then they verified 
multiplicative fusion of the spatial and temporal streams provided performance 
boost over 
an additive formulation in \cite{feichtenhofer2017spatiotemporal}. We 
re-implement the IF-TTN with fusion module in 
\cite{feichtenhofer2016spatiotemporal} and 
\cite{feichtenhofer2017spatiotemporal} and the experiment results are shown in 
Table \ref{table_IFM_com}. Experiment results show that our 
IFM performs much better than the fusion type used in 
\cite{feichtenhofer2016spatiotemporal} in 
\cite{feichtenhofer2017spatiotemporal}.

\begin{table}[t]
	\caption{Accuracy and inference speed comparison. The unit of inference 
		speed is the 
		fps. Experiments are conducted on UCF101 
		all splits. }
	\label{table_MV_com}
	\footnotesize
	\centering
	\begin{tabular}{c|ccccc|c|ccccc}
		\multicolumn{1}{c|}{Method} & \multicolumn{1}{c}{speed} & 
		\multicolumn{1}{c}{acc.} \\
		\hline
		\multicolumn{1}{c|}{Two-Stream I3D\cite{carreira2017quo}} & 
		\multicolumn{1}{c}{14} & \multicolumn{1}{c}{93.4}   \\
		\multicolumn{1}{c|}{TSN(RGB+Optical flow)\cite{wang2016temporal}} & 
		\multicolumn{1}{c}{14} & \multicolumn{1}{c}{94.0}   \\
		\hline
		\multicolumn{1}{c|}{DIN\cite{bilen2016dynamic}} & 
		\multicolumn{1}{c}{131} & \multicolumn{1}{c}{76.9}  \\
		\multicolumn{1}{c|}{C3D\cite{tran2015learning}} & 
		\multicolumn{1}{c}{314} & \multicolumn{1}{c}{82.3}  \\
		\multicolumn{1}{c|}{TSN(RGB)\cite{wang2016temporal}} & 
		\multicolumn{1}{c}{680} & \multicolumn{1}{c}{85.5}  \\
		\multicolumn{1}{c|}{TSN(RGB+RGB Difference)\cite{wang2016temporal}} & 
		\multicolumn{1}{c}{340} & \multicolumn{1}{c}{91.0}  \\
		\multicolumn{1}{c|}{RGB+EMV-CNN \cite{zhang2016real}} & 
		\multicolumn{1}{c}{390} & \multicolumn{1}{c}{86.4}  \\
		\multicolumn{1}{c|}{CoViAR\cite{wu2017compressed} } & 
		\multicolumn{1}{c}{240} & \multicolumn{1}{c}{90.4}  \\
		
		\multicolumn{1}{c|}{OFF\cite{sun2017optical} } & 
		\multicolumn{1}{c}{206} & \multicolumn{1}{c}{93.3}  \\
		\hline
		\multicolumn{1}{c|}{MV-IF-TTN} & 
		\multicolumn{1}{c}{142} & \multicolumn{1}{c}{94.5}  \\
	\end{tabular}
\end{table}
\textbf{Study on TTN:} 
We report the experiment results of each network branch in Table 
\ref{table_TTN}. TTN branch indicates that the predictions are made without 
ensembling classification scores of the 
Two-Stream CNN.
All these experiments are carried out with TSN framework. From Table 
\ref{table_TTN}, we can 
conclude that TTN is complementary to Two-Stream CNN and improves the accuracy 
by 1.9\% when combined.

\begin{figure}[t]
	\centering
	\includegraphics[width=0.47\textwidth]{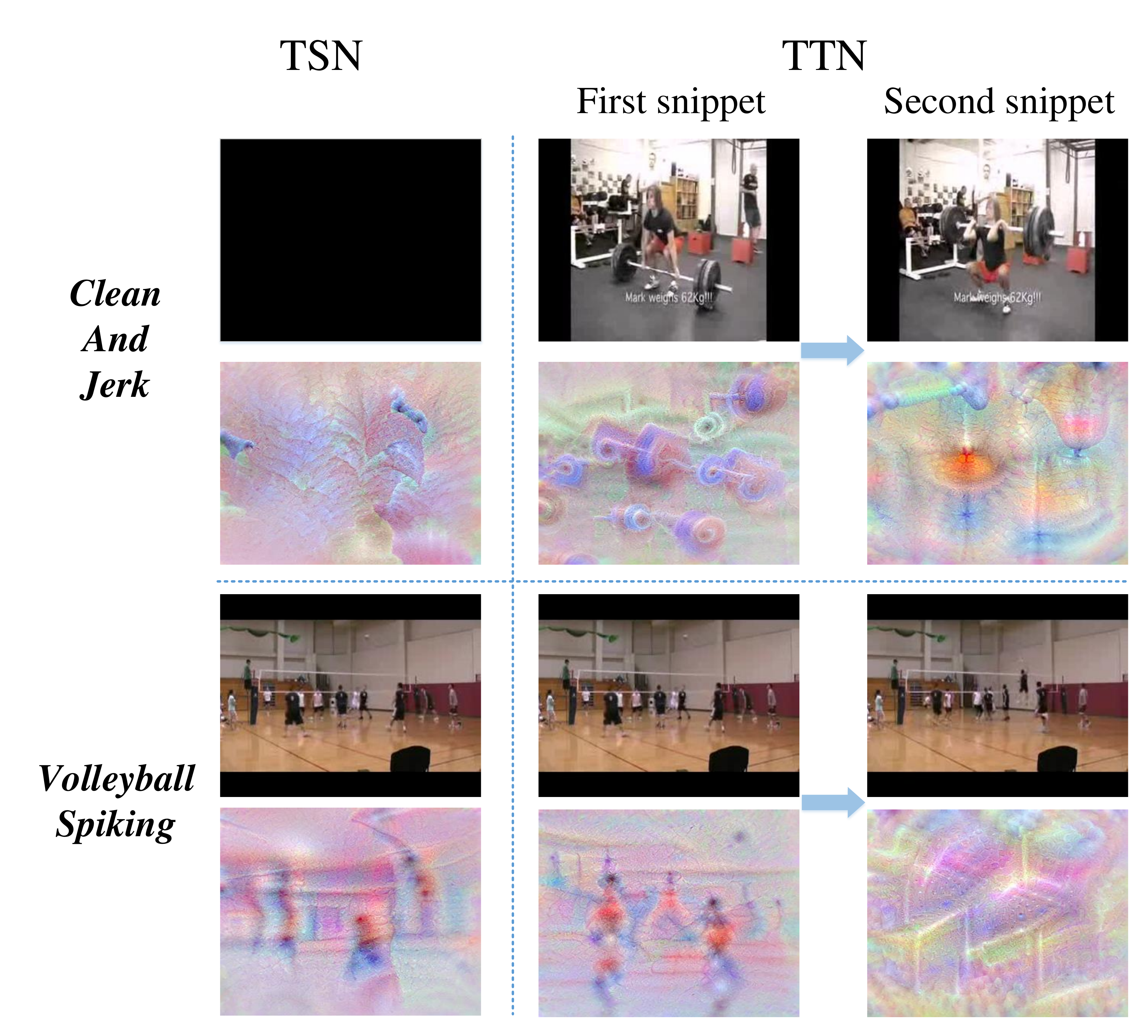}
	\caption{Class visualization of TSN and TTN using DeepDraw on action 
		categories:  ``\textit{CleanAndJerk}'' and 
		``\textit{VolleyballSpiking}''. The images are arranged in the same 
		way as in Figure \ref{fig_first}. The black image indicates that there 
		is no obvious corresponding RGB image to the generated one.}
	\label{sec_vis}
\end{figure}

Does TTN really learn the order relationship? To verify this, we adopt the 
DeepDraw \cite{deepdraw} toolbox to visualize our TTN models. This tool 
conducts iterative gradient ascent on input images with only white noises, and 
output class visualization based solely on class knowledge inside the CNN model 
after a number of iterations.  Since Our TTN takes two adjacent snippets as 
inputs, we adapt DeepDraw to deal with two inputs. The visualization results of 
TSN and our TTN are shown in Figure \ref{fig_first} and \ref{sec_vis}. From the 
visualization 
results, we can observe that TTN indeed learn the temporal transformations 
between two ordered video frames from adjacent segments. 
Taking  
``\textit{HighJump}'' for example, TTN models the transformation 
between 
human running in front of a high jump crossbar and human falling on the 
mat after skipping the crossbar while TSN mainly replies on the scene, 
e.g., the high jumping mat.

\textbf{Study on discriminate feature learning:} We study whether IF-TTN has 
learned the discriminate spatiotemporal features by 
visualizing the class-specific discriminate regions. The class-specific 
discriminative regions can be derived from classification network using Class 
Activation Maps (CAM) method \cite{zhou2016learning}. We 
visualized class-specific discriminative regions of spatial network, temporal 
network and IF-TTN, and show results in Figure \ref{fig_vis_ifm}. We 
can observe that 
spatial network mainly focuses the scene information, temporal network focuses 
on the short-term motion associated to that snippet, while IF-TTN covers the 
object regions and captures their motion track between the two snippets.

\textbf{Study on motion representation:} We replace optical flow with motion 
vectors as temporal stream inputs and evaluate the performance of motion vector 
based IF-TTN.

As shown in Table \ref{table_MV} the IF-TTN using motion vectors has 0.6\% 
degradation in recognition performance compared with flow-based IF-TTN.
The comparison with other real-time methods are provided in Table 
\ref{table_MV_com}. DIN represents the Dynamic Image Network proposed in 
\cite{bilen2016dynamic}. TSN (RGB),  TSN (RGB+RGB Difference) are from 
\cite{wang2016temporal}. OFF denotes the optical flow guided features in 
\cite{sun2017optical}. We denote our real-time IF-TTN as MV-IF-TTN. In order to 
conduct a more convincing 
comparison, we include two state-of-the-art optical flow based Two-Stream 
CNNs in the table. Our inference speed is tested on a single-core CPU (Intel 
Core i7-6850K) and a GeForce GTX 1080Ti GPU. From Table \ref{table_MV_com}, 
when replacing optical flow with motion vectors as motion inputs, IF-TTN 
achieves very competitive performance 94.5\% on UCF-101 dataset. This 
performance is even slightly higher than optical flow based Two-Stream TSN 
\cite{wang2016temporal} while the inference speed achieves 142 fps, which 
is about 10x faster than TSN.

Experiment results prove that our network is highly tolerant to the 
quality of motion input thanks to the combination of short-term 
spatiotemporal feature fusion, sequentially middle-term temporal modeling 
and 
long-term temporal consensus. EMV-CNN and CoViAR 
\cite{zhang2016real,wu2017compressed} also used motion 
vectors but the simple replacement without consideration of more effective 
spatiotemporal 
representation results in  a \textit{significant} 
performance degradation than optical-flow-based Two-Stream CNN.

\subsection{Comparison with the state of the art}\label{Eval_soa}
In this subsection, we compare IF-TTN with the state of the art. All experiment 
results are evaluated on HMDB-51 and UCF-101 over 
all three splits and shown in Table 
\ref{table_SoA}

The upper part of Table \ref{table_SoA} shows non-real-time methods while the 
lower part presents real-time methods. Notice that for non-real-time methods we 
assemble the optical flow and motion vectors based IF-TTN scores to make final 
predictions (denoted as Full IF-TTN).

We compare our method with both traditional approaches, like iDT 
\cite{wang2013action},  and deep learning 
based methods, such as Two-Stream CNN\cite{simonyan2014two}, C3D 
\cite{tran2015learning}, TSN 
\cite{wang2016temporal}, Temporal Deep convolutional Descriptors (TDD) 
\cite{wang2015action}, 
Long-term Temporal CNN \cite{varol2018long}, Spatiotemporal Pyramid Network 
\cite{wang2017spatiotemporal}, SaptioTemporal 
Multiplier Network \cite{feichtenhofer2017spatiotemporal},  Spatiotemporal 
Vector of Locally Max Pooled Features (ST-VLMPF) \cite{duta2017spatio}, Lattice 
LSTM \cite{sun2017lattice},
and Inflated 3D CNN (I3D) \cite{carreira2017quo} and Optical Flow guided 
Features (OFF) \cite{sun2017optical}. 
Our full IF-TTN achieves state-of-the-art results on both datasets. 
It is especially worth noting that the performance of MV-IF-TTN 
significantly outperforms the previous real-time methods.
\begin{table}[t]
	\caption{Comparison with state-of-the-art results. Experiments are 
		conducted on UCF-101 and HMDB-51
		over all three splits. '-' represents that the paper did not report the 
		corresponding result. }
	\label{table_SoA}
	\footnotesize
	\centering
	\begin{tabular}{c|c|c|c}
		\multicolumn{1}{c|}{Method} & \multicolumn{1}{c}{UCF-101} & 
		\multicolumn{1}{c}{HMDB-51} \\
		
		\hline
		
		\multicolumn{1}{c|}{iDT\cite{wang2013action}} & 
		\multicolumn{1}{c}{86.4} & \multicolumn{1}{c}{61.7}   \\
		
		\multicolumn{1}{c|}{Two stream CNN\cite{simonyan2014two}} & 
		\multicolumn{1}{c}{88.0} & \multicolumn{1}{c}{59.4}   \\
		\multicolumn{1}{c|}{TDD \cite{wang2015action}} & 
		\multicolumn{1}{c}{91.5} & \multicolumn{1}{c}{65.9}  \\
		\multicolumn{1}{c|}{Long Term Convolution \cite{varol2018long}} & 
		\multicolumn{1}{c}{91.7} & \multicolumn{1}{c}{64.8}  \\
		\multicolumn{1}{c|}{Spatiotemporal 
			Pyramid Network\cite{wang2017spatiotemporal}} & 
		\multicolumn{1}{c}{94.6} & \multicolumn{1}{c}{68.9}   \\
		
		\multicolumn{1}{c|}{Spatiotemporal Multiplier 
			Network\cite{feichtenhofer2017spatiotemporal}} & 
		\multicolumn{1}{c}{94.2} & \multicolumn{1}{c}{68.9}   \\
		
		\multicolumn{1}{c|}{Two stream TSN\cite{wang2016temporal}} & 
		\multicolumn{1}{c}{94.0} & \multicolumn{1}{c}{68.5}   \\
		\multicolumn{1}{c|}{ST-VLMPF\cite{duta2017spatio}} & 
		\multicolumn{1}{c}{93.6} & \multicolumn{1}{c}{69.5}   \\
		
		\multicolumn{1}{c|}{Two-Stream I3D\cite{carreira2017quo}}
		& 
		\multicolumn{1}{c}{93.4} & \multicolumn{1}{c}{66.4}   \\
		
		\multicolumn{1}{c|}{Lattice LSTM\cite{sun2017lattice}}& 
		\multicolumn{1}{c}{93.6} & \multicolumn{1}{c}{66.2}   \\
		\multicolumn{1}{c|}{Full OFF\cite{sun2017optical}} & 
		\multicolumn{1}{c}{96.0} & \multicolumn{1}{c}{74.2}  \\
		\multicolumn{1}{c|}{\textbf{Full IF-TTN}} & 
		\multicolumn{1}{c}{\textbf{96.2}} & \multicolumn{1}{c}{\textbf{74.8}}  
		\\
		\hline
		\multicolumn{1}{c|}{C3D\cite{tran2015learning}} & 
		\multicolumn{1}{c}{82.3} & \multicolumn{1}{c}{-}  \\
		\multicolumn{1}{c|}{TSN(RGB)\cite{wang2016temporal}} & 
		\multicolumn{1}{c}{85.7} & \multicolumn{1}{c}{51.0}  \\
		\multicolumn{1}{c|}{TSN(RGB+RGB Difference)\cite{wang2016temporal}} & 
		\multicolumn{1}{c}{91.0} & \multicolumn{1}{c}{-}  \\
		\multicolumn{1}{c|}{RGB+EMV-CNN } & 
		\multicolumn{1}{c}{86.4} & \multicolumn{1}{c}{53.0}  \\
		\multicolumn{1}{c|}{CoViAR\cite{wu2017compressed} } & 
		\multicolumn{1}{c}{90.4} & \multicolumn{1}{c}{59.1}  \\
		
		\multicolumn{1}{c|}{real-time OFF\cite{sun2017optical} } & 
		\multicolumn{1}{c}{93.3} & \multicolumn{1}{c}{-}  \\
		\multicolumn{1}{c|}{\textbf{MV-IF-TTN}} & 
		\multicolumn{1}{c}{\textbf{94.5}} & \multicolumn{1}{c}{\textbf{70.0}}  
		\\
	\end{tabular}
\end{table}


\section{Conclusion}
In this paper, we have proposed the IF-TTN to learn discriminate 
spatiotemporal 
features 
for video action recognition. Specially, the IFM is designed to fuse the 
appearance and motion features at multiple spatial 
scales for each video snippet, and the TTN is 
employed to model the middle-term temporal transformation between the 
neighboring snippets. Our network achieves the state-of-the-art results on two 
most popular action recognition datasets. The real-time version of IFM-TTN 
implemented on motion vectors achieves significant improvement against the 
state-of-the-art real-time methods.

{\small
\bibliographystyle{ieee}
\bibliography{action}
}

\end{document}